\documentclass{article} 
\usepackage{iclr2027_conference,times}

\usepackage{amsmath,amsfonts,bm}

\def\eqref#1{equation~\ref{#1}}

\def\1{\bm{1}}

\def\va{{\bm{a}}}

\DeclareMathAlphabet{\mathsfit}{\encodingdefault}{\sfdefault}{m}{sl}
\SetMathAlphabet{\mathsfit}{bold}{\encodingdefault}{\sfdefault}{bx}{n}

\usepackage{hyperref}
\usepackage{url}

\usepackage[utf8]{inputenc}
\usepackage[T1]{fontenc}
\usepackage{hyperref}
\usepackage{url}
\usepackage{booktabs}
\usepackage{amsfonts}
\usepackage{amsmath}
\usepackage{nicefrac}
\usepackage{microtype}
\usepackage{xcolor}
\usepackage{graphicx}
\usepackage{longtable}
\usepackage{amssymb}
\usepackage{array}
\usepackage{svg}
\usepackage{makecell}
\usepackage{multirow}
\usepackage{caption}
\usepackage{xcolor}
\usepackage{bbold}
\usepackage{subcaption}
\usepackage{anyfontsize} 

\newcommand{\myparagraph}[1]{\textbf{#1}\quad}

\newcommand{\good}[1]{Good}
\newcommand{\neutral}[1]{Neutral}
\newcommand{\bad}[1]{Bad}
\newcommand{\yes}[1]{\includegraphics[width=1em,height=1em]{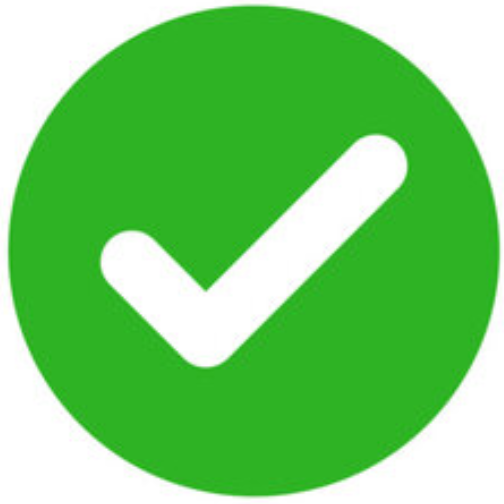}}
\newcommand{\no}[1]{\includegraphics[width=1em,height=1em]{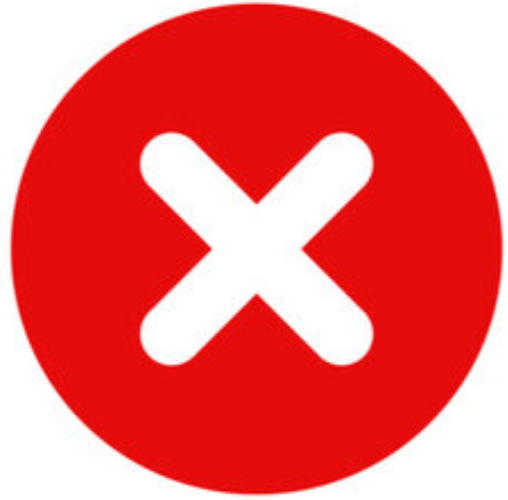}}
\newcommand{\maybe}[1]{\includegraphics[width=1em,height=1em]{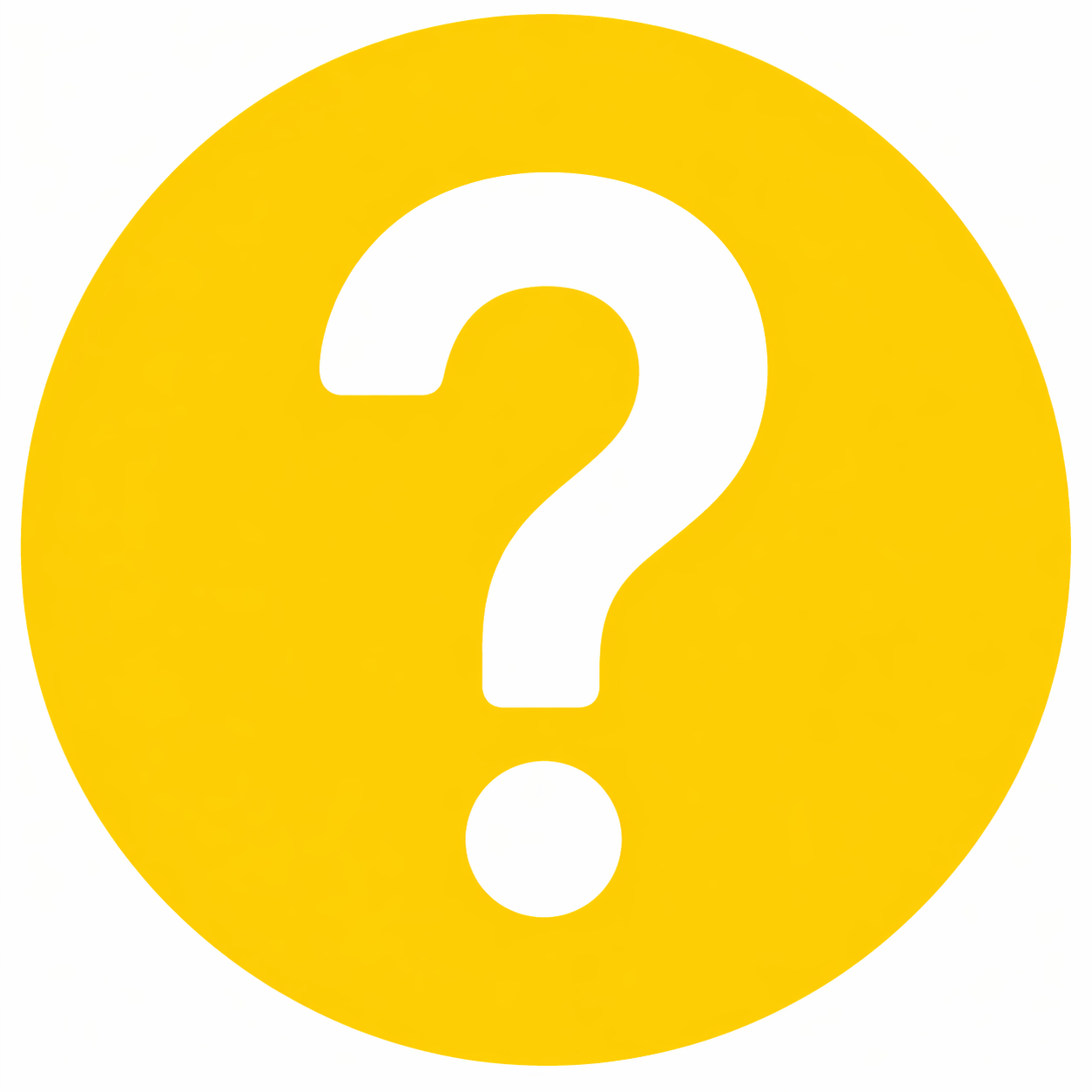}}
\newcommand{\NA}[1]{/}

\newcommand{\environment}[1]{\includegraphics[width=1em,height=1em]{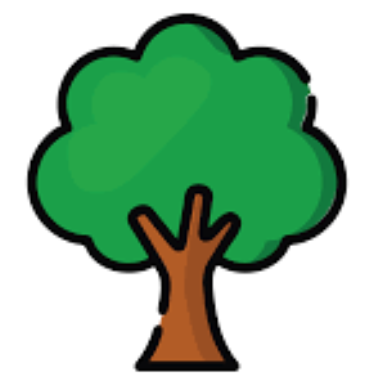}}
\newcommand{\science}[1]{\includegraphics[width=1em,height=1em]{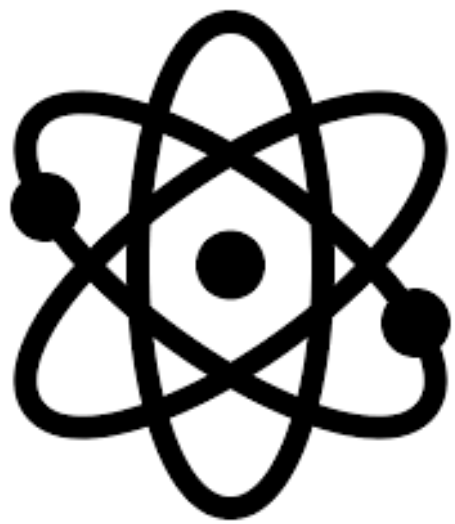}}
\newcommand{\finance}[1]{\includegraphics[width=1em,height=1em]{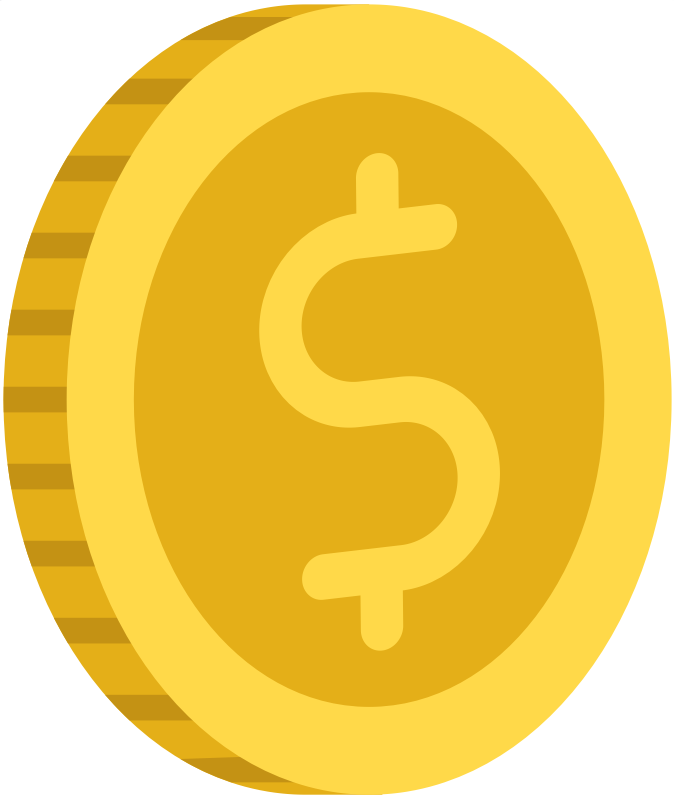}}
\newcommand{\medicine}[1]{\includegraphics[width=1em,height=1em]{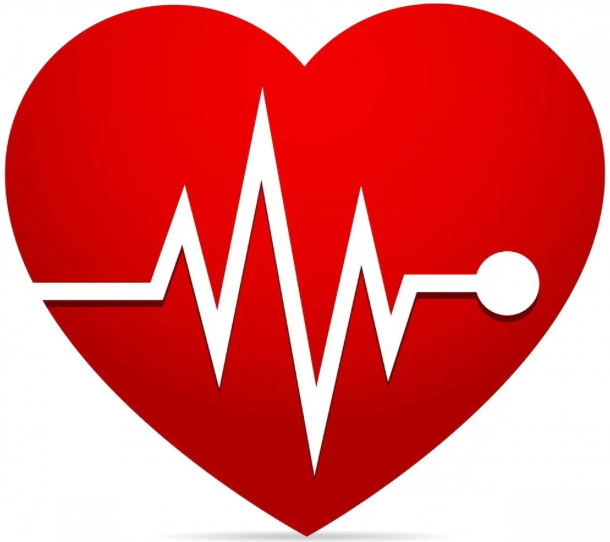}}
\newcommand{\technology}[1]{\includegraphics[width=1em,height=1em]{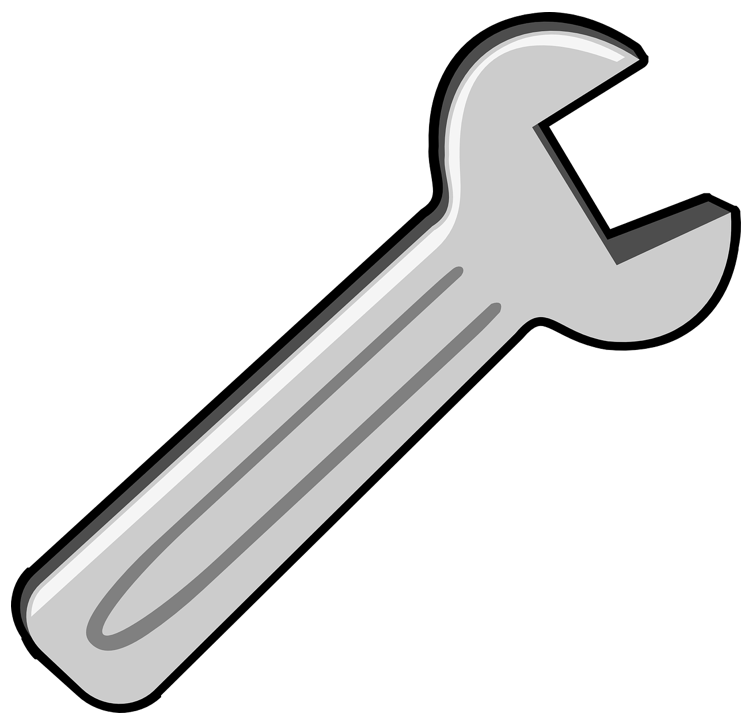}}
\newcommand{\synthetic}[1]{\includegraphics[width=1em,height=1em]{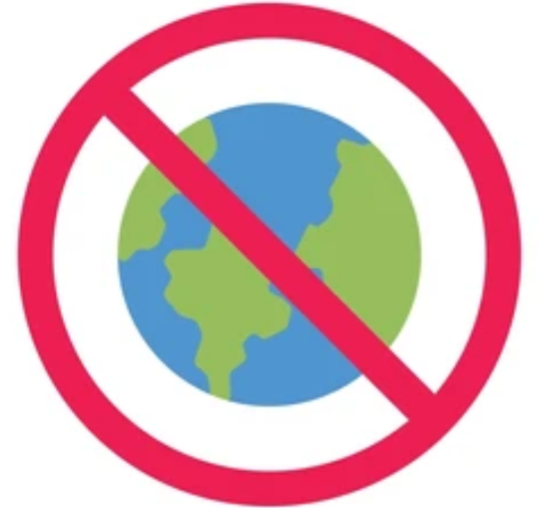}}

\newcommand{\benchmarkName}[1]{SelectArena}
\newcommand{\numberOfDatasets}[1]{71}
\newcommand{\numberOfMethods}[1]{15}
\newcommand{\numberOfModels}[1]{5}
\newcommand{\numberOfRepeats}[1]{3}
\newcommand{\numberOfFolds}[1]{10}
\newcommand{\numberOfSplits}[1]{5}

\usepackage[nameinlink]{cleveref}

\usepackage[page]{appendix}
\usepackage{etoolbox}

\renewcommand{\appendixtocname}{Table of Contents.}

\makeatletter
\let\oldappendix\appendices

\renewcommand{\appendices}{%
  \clearpage
  \renewcommand{\thesection}{\Roman{section}}
  \let\tf@toc\tf@app
  \addtocontents{app}{\protect\setcounter{tocdepth}{2}}
  \immediate\write\@auxout{%
    \string\let\string\tf@toc\string\tf@app^^J
  }
  \oldappendix
}%

\newcommand{\listofappendices}{%
  \begingroup
  \renewcommand{\contentsname}{\appendixtocname}
  \let\@oldstarttoc\@starttoc
  \def\@starttoc##1{\@oldstarttoc{app}}
  \tableofcontents
  \endgroup
}

\title{\benchmarkName{}: A Modern Benchmark for Feature Selection on Real-world Tabular Data}
\iclrfinalcopy
\author{
  \textbf{Bastian Schäfer}$^{1, *}$ \quad
  \textbf{Dominika Matus}$^{2, *}$ \quad
  \textbf{Lennart Purucker}$^{3, 2}$ \quad
  \textbf{Ricardo Knauer}$^4$ \\
  \quad \textbf{Erik Rodner}$^4$ \quad\quad
  \textbf{Frank Hutter}$^{3,5,2}$ \quad\quad
  \textbf{Anton Sokolov}$^1$ \quad\quad
  \textbf{Arnaud Liefooghe}$^1$ \\
  \quad\quad $^1$Université du Littoral-Côte-d'Opale \quad $^2$University of  Freiburg
  \quad $^3$Prior Labs \\
  \quad\quad\quad $^4$HTW Berlin \quad\quad $^5$ELLIS Institute Tübingen \quad\quad $^*$Equal contribution. \\
  \texttt{bastian.schafer@univ-littoral.fr, matusd@cs.uni-freiburg.de}
}

\begin{document}
\maketitle

\begin{abstract}
Selecting a subset of features (columns) from a tabular dataset is a ubiquitous challenge in real-world applications. 
Industry experts seek to minimize model training costs, climate scientists require an interpretable subset of factors to understand extreme weather events, and clinicians need to choose among diagnostic tests due to financial and time constraints. 
When looking at the scientific literature to determine which feature selection method to use, practitioners across domains face an unrepresentative landscape of benchmarking.
Despite extensive research, existing feature selection benchmarks provide limited guidance because they rely on outdated models, ad hoc evaluation protocols, or artificially generated data that do not reflect real-world problems. 
To close the gap between scientific benchmarking and real-world applications, we introduce \benchmarkName{}, a modern benchmark for feature selection.
We compare \numberOfMethods{} widely used feature selection methods on \numberOfDatasets{} manually curated real-world tabular datasets, covering classification and regression problems, using standardized evaluation protocols and state-of-the-art tabular (foundation) models.
Our results reveal clear trade-offs across predictive performance, runtime, stability, and validity. Tree-based methods, CART and RFImportance, along with F-Test are Pareto-optimal when taking into account predictive performance and runtime.
F-Test, information-theory-, and distance-based methods lead on stability and validity, with F-Test emerging as a balanced choice across our metrics.
Search-based methods are slow, unstable and achieve low performance.
\benchmarkName{} turns these findings into practical guidance and, as an extensible platform, supports future feature selection research.
\end{abstract}

\begin{figure}[h!]
  \centering
  \includegraphics[width=\textwidth]{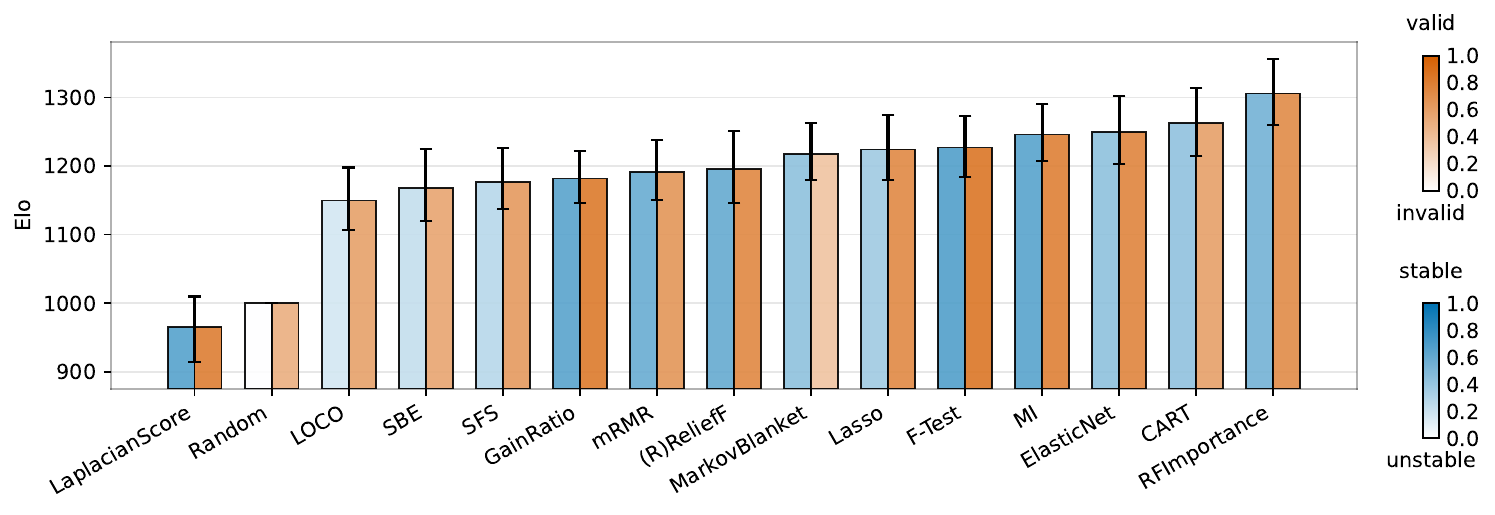}
  \caption{\textbf{\benchmarkName{} at a glance.} We present Elo ratings of the $\numberOfMethods{}$ feature selection methods ordered left to right from worst- to best-performing method. Elo rating summarizes how often a method beats another on the same dataset, budget, and downstream model, with random selection anchored at $1000$ (Appendix~\ref{par:elo-rating}). Error bars are $95\%$ confidence intervals from $1000$ cluster bootstrap resamples over the $\numberOfDatasets{}$ datasets. The left bar of each pair is shaded by \emph{stability} (how much the selected subset changes across resamples) and the right bar by \emph{validity} (how well injected noise features are rejected), where darker is better for both.}
  \label{figure-results_performance-elo}
\end{figure}

\section{Introduction}\label{section-introduction}
Tabular data dominates many medical, industrial, and scientific applications~\citep{borisov-nnls22a, erickson2025tabarena}. However, collecting such data is often costly and requires labor-intensive measurements, expensive instruments, or access to specialized facilities~\citep{hedou2024stabllasso, wang2025towards}. At the same time, the growing dimensionality of datasets decreases interpretability, increases model training and inference costs, and forces practitioners to trade off performance against computation time. 

Feature selection addresses these challenges by identifying a subset of the most informative features, thereby reducing data collection and training costs while improving transparency and interpretability. Early work~\citep{guyon2003introduction} has framed the objectives of feature selection around predictive performance, computational cost, and interpretability. Later work has added stability~\citep{nogueira2016stability} and validity~\citep{guyon2007benchmark} as standard criteria for reliable feature selection. Within this problem formulation, hundreds of methods have been developed ~\citep{li2018scikit_feature_article}, each making different trade-offs across these metrics. The benchmarking literature, however, remains highly fragmented and despite its long history \citep{jain1997FSbenchmark,kohavi1997wrappers,reunanen2003insufficientBenchmarkFS}, does not provide sufficient guidance to practitioners.

Existing benchmarks have not kept pace with the development of modern tabular models, in particular foundation models, and rely on unrealistic evaluation protocols. Most use small-scale comparisons on synthetic or limited real-world datasets with $\leq$ 10 methods and $\leq$ 20 real-world datasets~\citep{bommert2020benchmarkFS, cherepanova2023benchmarkFS, bakhmach2026roofs} (Section~\ref{section-related-work}, Table~\ref{table-related-work-benchmarks}). To address these gaps, we introduce \benchmarkName{}, a benchmark of \numberOfMethods{} feature selection methods evaluated on \numberOfDatasets{} manually curated real-world tabular datasets across predictive performance, runtime, stability, and validity, using state-of-the-art tabular (foundation) models. Our contributions can be summarized as follows:
\begin{itemize}
    \item We analyze 368 datasets from the feature selection and machine learning literature and manually curate a set of \numberOfDatasets{} real-world tabular datasets following a strict data curation protocol. We release the resulting datasets.
    \item We review 296 feature selection methods and select a set of \numberOfMethods{} methods according to a strict method curation protocol. Where well-maintained codebases are unavailable, we provide our own implementations.
    \item We provide a standardized evaluation pipeline covering predictive
    performance, stability, and validity, alongside runtime, and evaluate all methods with five downstream models including state-of-the-art tabular foundation model. We share our code to make our pipeline open-source.
\end{itemize}

Our results reveal clear trade-offs across predictive performance, runtime, stability, and validity.  Tree-based (RFImportance, CART) and regularization-based methods (ElasticNet, Lasso) achieve the highest predictive performance, while F-Test and GainRatio are the fastest and, alongside MI and LaplacianScore, lead on stability and validity. 
With all four criteria taken into account, F-Test emerges as the most balanced choice across them. 
Search-based methods are the counterexample: they are slow and rank near the bottom on performance and stability. 
However, no method wins on every criterion, so the right choice still depends on which objective a practitioner prioritizes. 
SelectArena turns these findings into practical guidance and, as an extensible open-source platform, supports future feature selection research.
\section{Related Work}\label{section-related-work}
Recent years have seen numerous feature selection benchmarks. Despite extensive research, practitioners still lack clear guidance on selecting the most suitable method for a given context. Existing benchmarks are typically limited in scope by examining only a small number of methods, most often no more than 10 \citep{bouchlaghem2022reviewFS, kohavi1997wrappers}, or by focusing exclusively on specific subgroups \citep{bommert2020benchmarkFS, hopf2021benchmarkFilterFS}. Moreover, many studies rely on unrepresentative or limited data sources, such as synthetic data \citep{bolon2013review, hopf2021benchmarkFilterFS}, domain-specific data \citep{bakhmach2026roofs, li2022specialBenchmarkFS, umar2025effects}, or non-tabular data (text or image) \citep{guyon2007benchmark, kou2020benchmarkFSText, miao2016surveyFS}. In most cases, the number of real-world tabular datasets does not exceed 20.

Furthermore, most benchmarks focus only on predictive performance \citep{chandrashekar2014surveyFS, reunanen2003insufficientBenchmarkFS, sheikhpour2017surveyFSsemiSupervised}. Few studies consider the stability of the selected subsets \citep{bommert2020benchmarkFS, brown2012ITFSframework, kou2020benchmarkFSText, saeys2008insufficientBenchmarkFS}, even fewer the validity \citep{bolon2013review, cherepanova2023benchmarkFS, guyon2007benchmark} or both \citep{bakhmach2026roofs, hopf2021benchmarkFilterFS}. In addition, even recent works predominantly rely on classical downstream models, with little consideration of state-of-the-art approaches like tuned gradient-boosted trees or pretrained tabular foundation models \citep{bakhmach2026roofs, cherepanova2023benchmarkFS, umar2025effects}.

In contrast, \benchmarkName{} comprehensively benchmarks a diverse set of feature selection methods across a large and broad dataset collection, evaluates the predictive performance as well as the stability and validity, and incorporates state-of-the-art models. Table \ref{table-related-work-benchmarks} provides a systematic comparison of existing feature selection benchmarks and \benchmarkName{} with respect to included methods, datasets, models, and evaluation metrics.

\begin{table}[ht]
\caption{\textbf{Comparison of feature selection benchmarks.}
	We present former benchmarks in comparison to \benchmarkName{}. 
    \textbf{\# Methods}: how many out of the total number of benchmark methods correspond to our first three curation criteria (not a variant of a method and applicable across tabular classification and regression tasks).
    \textbf{\# Datasets}: how many out of the total number of benchmark datasets satisfy our curation criteria for high-quality tabular data (Section \ref{subsection-benchmarking-method-curation}). 
    \textbf{Domain}: visualizes the dataset domain: \synthetic{} (synthetic), \environment{} (environmental), \medicine{} (biomedical), \technology{} (technological), \science{} (scientific), \finance{} (financial) or \no{} (not applicable). 
    \textbf{SotA models}: whether the benchmark includes state-of-the-art downstream models, such as tuned gradient-boosted trees or pretrained tabular foundation models \citep{erickson2025tabarena}, with the options \yes{} (yes) or \no{} (no). 
    The final three columns (\textbf{Perf}, \textbf{Stab}, \textbf{Val}) indicate whether the benchmark reports an evaluation of the predictive performance, stability, or validity.}
\label{table-related-work-benchmarks}
\centering
\small
\begin{tabular}{lccccccc}
    \toprule
    \textbf{Benchmark} & \makecell{\textbf{\# Meth-}\\\textbf{ods}} & \makecell{\textbf{\# Data-}\\\textbf{sets}}
            & \textbf{Domain} & \makecell{\textbf{SotA}\\\textbf{models}} & \textbf{Perf} & \textbf{Stab} & \textbf{Val} \\
    \midrule
   \cite{kohavi1997wrappers} & \textcolor{red}{3}/4 & \textcolor{red}{5}/14  & \synthetic{}\medicine{}\finance{} & \no{} & \yes{} & \no{} & \no{} \\
    \cite{jain1997FSbenchmark} & \textcolor{red}{8}/15 & \textcolor{red}{0}/1  & \technology{} & \no{} & \yes{} & \no{} & \no{} \\
    \cite{reunanen2003insufficientBenchmarkFS} & \textcolor{red}{2}/2  & \textcolor{red}{2}/7  & \environment{}\medicine{}\technology{} & \no{} & \yes{} & \no{} & \no{} \\
    \cite{guyon2007benchmark} & \textcolor{red}{1}/3 & \textcolor{red}{0}/5  & \no{} & \no{} & \yes{} & \no{} & \yes{} \\
    \cite{saeys2008insufficientBenchmarkFS} & \textcolor{red}{4}/4 & \textcolor{red}{6}/6  & \medicine{} & \no{} & \yes{} & \yes{} & \no{} \\
    \cite{brown2012ITFSframework} & \textcolor{red}{7}/9 & \textcolor{red}{4}/15  & \environment{}\medicine{}\technology{} & \no{} & \yes{} & \yes{} & \no{} \\
    \cite{bolon2013review} & \textcolor{red}{7}/11 & \textcolor{red}{0}/11  & \synthetic{} & \no{} & \yes{} & \no{} & \yes{} \\
    \cite{chandrashekar2014surveyFS} & \textcolor{red}{3}/4 & \textcolor{red}{2}/6 & \medicine{}\technology{} & \no{} & \yes{} & \no{} & \no{} \\
    \cite{miao2016surveyFS} & \textcolor{red}{8}/8 & \textcolor{red}{1}/12  & \no{} & \no{} & \yes{} & \no{} & \no{} \\
    \cite{sheikhpour2017surveyFSsemiSupervised} & \textcolor{red}{3}/9 & \textcolor{red}{0}/3  & \medicine{} & \no{} & \yes{} & \no{} & \no{} \\
    \cite{li2018scikit_feature_article} & \textcolor{green}{18}/45 & \textcolor{red}{7}/25  & \medicine{}{} & \no{} & \no{} & \no{} & \no{} \\
    \cite{kou2020benchmarkFSText} & \textcolor{red}{1}/10 & \textcolor{red}{0}/10  & \no{}{} & \no{} & \yes{} & \yes{} & \no{} \\
    \cite{bommert2020benchmarkFS} & \textcolor{red}{12}/22 & \textcolor{red}{1}/16 & \environment{}\medicine{}\technology{}\finance{} & \no{} & \yes{} & \yes{} & \no{} \\
    \cite{hopf2021benchmarkFilterFS} & \textcolor{red}{12}/58 & \textcolor{red}{0}/13 & \synthetic{} & \no{} & \yes{} & \yes{} & \yes{} \\
    \cite{bouchlaghem2022reviewFS} & \textcolor{red}{5}/5 & \textcolor{red}{0}/7  & \technology{} & \no{} & \yes{} & \no{} & \no{} \\
    \cite{li2022specialBenchmarkFS} & \textcolor{red}{7}/8 & \textcolor{red}{0}/15  & \medicine{} & \no{} & \yes{} & \no{} & \no{} \\
   \cite{cherepanova2023benchmarkFS} & \textcolor{red}{8}/9 & \textcolor{red}{1}/12  & \technology{}\science{}\finance{} & \no{} & \yes{} & \no{} & \yes{} \\
    \cite{schutz2025afabench} & \textcolor{red}{8}/10 & \textcolor{red}{1}/11  & \synthetic{}\medicine{}\technology{}\finance{} & \no{} & \yes{} & \no{} & \no{} \\
    \cite{umar2025effects} & \textcolor{red}{1}/1 & \textcolor{red}{0}/3  & \technology{}{} & \no{} & \yes{} & \no{} & \no{} \\
    \cite{bakhmach2026roofs} & \textcolor{red}{14}/22 & \textcolor{red}{0}/1 & \medicine{} & \no{} & \yes{} & \yes{} & \yes{} \\
    \textbf{SelectArena (Ours)} & 15/15 & \textbf{\textcolor{green}{\numberOfDatasets{}}}/\numberOfDatasets{}  & \environment{}\medicine{}\technology{}\science{}\finance{} & \yes{} & \yes{} & \yes{} & \yes{} \\
    \bottomrule
\end{tabular}
\end{table}
\section{SelectArena}\label{section-benchmarking}
\subsection{Problem Setup}
\label{section-problem-setup}

In practical applications, feature acquisition is often constrained by limited resources such as time, financial cost, or computational overhead. 
As a result, only a subset of all available features can be collected or processed, making it essential to select the most informative features under a given budget~$b$. 
This setting can be formalized as follows: 
Let $\mathcal{F} = \{1,\dots,p\}$ denote the full feature set and let $S \subseteq \mathcal{F}$ be a selected subset with associated cost $c(S) = \sum_{i \in S} c_i$. 
We define a model-dependent subset utility function $u_{M}: 2^{\mathcal{F}} \to \mathbb{R}$, which evaluates the utility of a feature subset $S$ under a given model $M$. 
The budget-constrained feature selection problem can then be formulated as:
$\max_{S \subseteq \mathcal{F}} u_{M}(S)$ subject to $c(S) \le b$.
In certain cases, this problem can be solved to optimality, for example via exhaustive enumeration when the feature (sub)set size is small or via best subset selection in some models \citep{tillmann2024cardinality}. 
In practice, however, exact computation is often infeasible~\citep{Amaldi1998Approximability}, and approximate (heuristic) solution methods are therefore more commonly employed. 
As a special case, when feature costs are uniform ($c_i = 1$) and the budget is fixed to the cardinality ($b = k$), the constraint reduces to $|S| \le k$, recovering the classical cardinality-constrained feature selection problem \citep{knauer2023cost}. 
While the budget-constrained formulation is more general, it may introduce a degree of arbitrariness in practice, as small changes in the budget can significantly alter $S$ without necessarily reflecting meaningful differences in resource availability. 
By contrast, cardinality constraints provide a simpler, more natural abstraction in many settings and are therefore mostly adopted in practice and used in \benchmarkName{}.

\subsection{Benchmarking Setup}\label{subsection-benchmarking-setup}

\begin{figure}[h]
  \centering
    \begin{minipage}[t]{0.48\textwidth}
    \centering
    \begin{minipage}[t]{0.48\textwidth}
      \includegraphics[width=\textwidth]{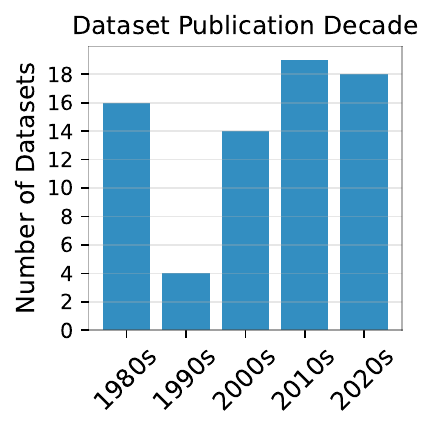}
    \end{minipage}
    \hfill 
    \begin{minipage}[t]{0.48\textwidth}
      \includegraphics[width=\textwidth]{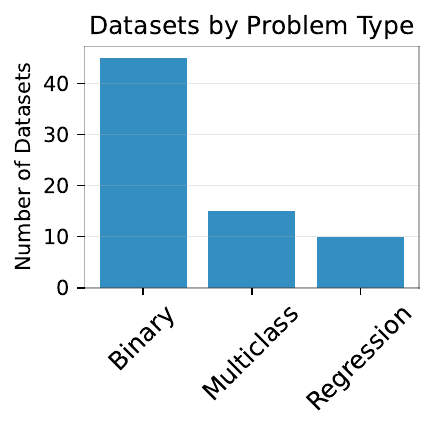}
    \end{minipage}   
    \begin{minipage}[t]{0.48\textwidth}
      \includegraphics[width=\textwidth]{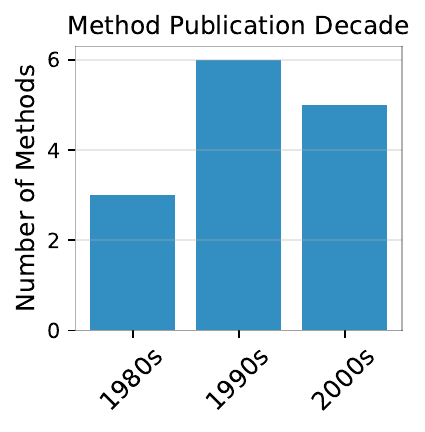}
    \end{minipage}
    \hfill
    \begin{minipage}[t]{0.48\textwidth}
      \includegraphics[width=\textwidth]{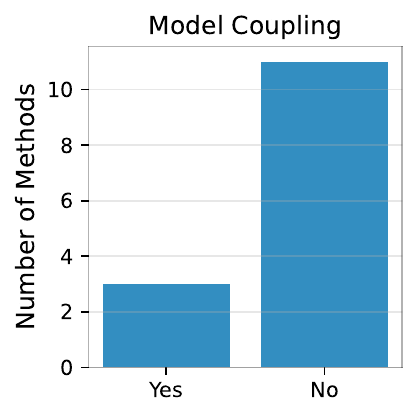}
    \end{minipage}
  \end{minipage}
  \begin{minipage}[d]{0.5\textwidth}
    \centering
    \vspace{0.2cm} 
    \includegraphics[width=\textwidth]{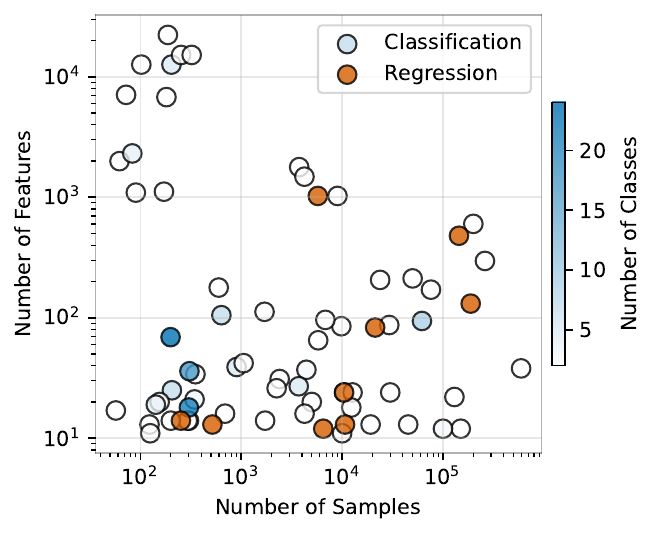}
  \end{minipage}
  \caption{\textbf{Method and dataset characteristics.} \benchmarkName{} represents a diverse collection of widely adopted feature selection methods as well as high-quality real-world datasets.}
  \label{figure-methodology-characterisitics}
\end{figure}

\myparagraph{Method Curation Protocol.}\label{subsection-benchmarking-method-curation}
We reviewed 33 prominent survey, benchmark, and related papers, covering seminal works and specialized contributions across the feature selection landscape, from foundational theory to domain-specific applications and recent benchmarking with full list of works in Table~\ref{appendix-table-surveys}. From this pool, we identified 296 unique feature selection methods (Table \ref{appendix-table-methods-overview}) and selected a representative set of methods based on the following criteria: (1) suitability for tabular data, (2) suitability for all tasks (binary classification, multi-class classification and regression), (3) methodological independence (not merely a variant of another method), (4) number of citations in the relevant literature, and (5) subgroup coverage. 
The \numberOfMethods{} feature selection methods that have been included in \benchmarkName{} reflect a mix of widely adopted and methodologically diverse approaches, with Table \ref{table-benchmarking-selected-methods} and Figure \ref{figure-methodology-characterisitics} summarizing their key characteristics. For methods requiring model coupling, we use LGBM.  More details on the curation criteria, curation insights, as well as an overview of all investigated methods, their source, the number of citations in the relevant literature, and the final decision on inclusion is available in Appendix \ref{appendix-section-methods}.

\begin{table}[ht]
\caption{\textbf{Benchmarked feature selection methods.} 
	We report all included methods in \benchmarkName{}. 
	\textbf{Determinism:} whether a method yields the same subset for the same data across repeated evaluations (also referred to as consistency in \citet{dembinsky2026unifying}). The options are \yes{} (yes), \no{} (no), or \maybe{} (model-dependent). \textbf{Model coupling}: whether feature selection can be performed in conjunction with the downstream model $M$ (related to portability in \citet{dembinsky2026unifying}).
    \textbf{Subgroup}: we specify which subgroup a method belongs to (Appendix \ref{section-taxonomy}). 
    \textbf{Citations}: reflects a method's popularity and impact within the relevant feature selection literature, measured as the frequency of occurrence across relevant survey and benchmark works (Section \ref{subsection-benchmarking-method-curation}).
    For implementation details, please refer to Appendix \ref{appendix-subsection-methods-implementation}.}
\label{table-benchmarking-selected-methods}
\centering
\small
\setlength{\tabcolsep}{4pt}
\begin{tabular}{lccccc}
    \toprule
    \textbf{Name} & \textbf{Deter-} & \textbf{Model} & \textbf{Subgroup} & \textbf{Cita-} & \textbf{Implemen-} \\
     & \textbf{minism} & \textbf{coupling} & & \textbf{titons} & \textbf{tation} \\
    \midrule
    CART \citep{breiman1984CART} & \yes{} & \no{} & Tree & 4 & scikit-learn \\
    ElasticNet \citep{zou2005elasticnet} & \no{} & \no{} & Regularization & 3 & scikit-learn \\
    F-test \citep{st1989anova} & \yes{} & \no{} & Correlation & 6 & scikit-learn \\
    GainRatio \citep{quinlan1986decistiontrees} & \yes{} & \no{} & Info theory & 3 & Us \\
    LaplacianScore \citep{he2005laplacianscore} & \no{} & \no{} & Distance & 5 & Us \\
    Lasso \citep{tibshirani1996lasso} & \no{} & \no{} & Regularization & 8 & scikit-learn  \\
    LOCO \citep{duch2006filter} & \maybe{} & \yes{} & Leave-one-out & 3 & Us \\
    MarkovBlanket \citep{koller1996MarkovBlanket} & \no{} & \no{} & Backward search & 3 & Us \\
    mRMR \citep{peng2005mRMR} & \yes{} & \no{} & Info theory & 12 & Us \\
    Mutual Information \citep{battiti1994MIM} & \yes{} & \no{} & Info theory & 6 & scikit-learn \\
    Random (Baseline) & \no{} & \no{} & \NA{} & \NA{} & Us \\
    (R)ReliefF \citep{kononenko1997reliefF} & \yes{} & \no{} & Distance & 10 & Us \\
    RFImportance \citep{breiman2001rf} & \no{} & \no{} & Tree & 8 & scikit-learn \\
    SBE \citep{guyon2002fsSVM} & \maybe{} & \yes{} & Backward search & 8 & Us \\
    SFS \citep{cover1977possible} & \maybe{} & \yes{} & Forward search & 10 & Us \\
   \bottomrule
\end{tabular}
\end{table}

Our method implementation builds on AutoGluon \citep{erickson-arxiv20a} and TabArena \citep{erickson2025tabarena}. We introduce a feature selection routine into TabArena that is integrated in the model-agnostic preprocessing pipeline of AutoGluon. The routine enforces a cardinality constraint $k$ (Section \ref{section-problem-setup}), a time limit and a fallback to random selection in case of timeout (Appendix \ref{par:fallback}) and custom preprocessing per method (Appendix \ref{appendix-subsection-methods-implementation}). To support time-bounded execution, the remaining runtime is periodically checked and intermediate results are returned when the limit is reached.

\begin{figure}[t]
  \centering
   \includegraphics[width=\textwidth]{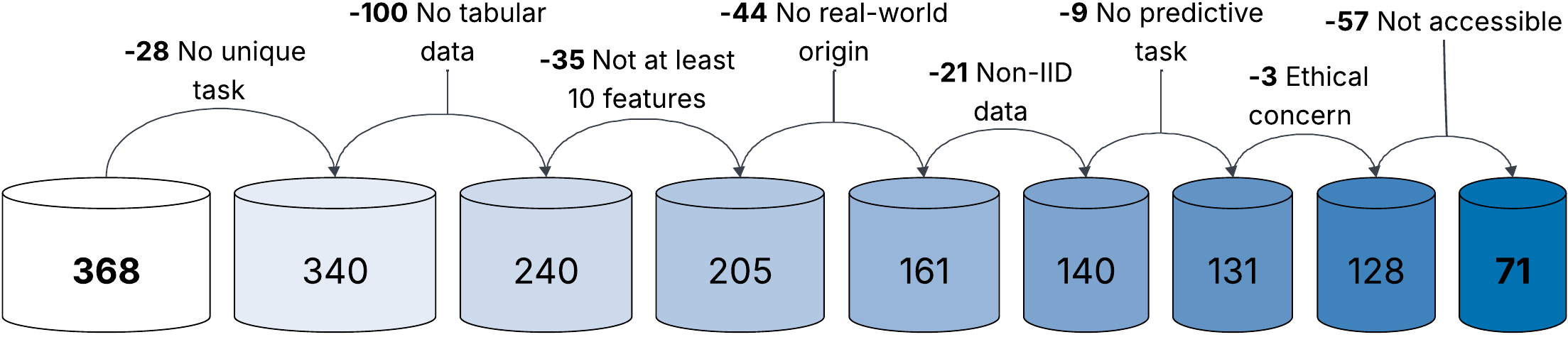}
  \caption{\textbf{Dataset curation process.} The reason and number of datasets excluded in each step of our dataset curation pipeline are shown, along with the remaining number of datasets after each step.}
  \label{figure-benchmarking-dataset-curation-process}
\end{figure}

\myparagraph{Dataset Curation Protocol.}\label{subsection-benchmarking-dataset-curation}
Real-world tabular datasets are the foundation of \benchmarkName{}, so we build on the curation standards of TabArena \citep{erickson2025tabarena} and extend its dataset collection towards feature selection. In order to do so, we surveyed the feature
selection survey and benchmark literature, the papers of our selected methods, and the related explainable AI literature, covering 73 articles in total (Appendix~\ref{appendix-subsection-dataset-sources}), from which we identified 368 candidate datasets. We then applied the following curation criteria: (1)~unique predictive task definition, (2)~tabular data (\textit{e.g.}, no text or image data), (3)~minimum of 10 features for meaningful selection, (4)~real-world origin (no synthetic data), (5)~IID data suitable for random splitting, (6)~genuine predictive task, (7)~no ethical concerns with the predictive task, and (8)~accessible source data. Implementing these criteria results in \numberOfDatasets{} high-quality real-world datasets in a curation process presented in Figure \ref{figure-benchmarking-dataset-curation-process}. The datasets span many domains (Table ~\ref{table-related-work-benchmarks}), cover diverse prediction tasks and differ in size as well as feature dimensionality (Figure \ref{figure-methodology-characterisitics}).  More details on the curation criteria, curation insights, and on overview of the selected datasets with their key properties can be found in Appendix \ref{appendix-section-datasets}.

\myparagraph{Downstream Models.}\label{subsection-benchmarking-models}
To evaluate the selected subset of features on the subsequent prediction tasks, we use five downstream models: linear model (Linear), random forest (RF)~\citep{breiman2001rf}, LightGBM (LGBM)~\citep{ke2017lightgbm}, RealMLP~\citep{holzmüller2025betterdefaultstrongpretuned} and TabICLv2 \citep{qu2026tabiclv2} spanning diverse inductive biases. We use AutoGluon's default configuration in our experiments. For LGBM, we conduct an ablation study on the impact of hyperparameter tuning (Appendix \ref{appendix-subsection-benchmarking-performance}). 

\myparagraph{Hardware usage.}\label{subsection-benchmarking-hardware}
CPU experiments (stability, validity and performance with the Linear, RF and LGBM model) ran on an AMD EPYC 9655 with 96 cores and GPU experiments (performance with RealMLP and TabICLv2) primarily on an NVIDIA L40 with 48\,GB VRAM and in rare cases on NVIDIA H200 141\,GB VRAM. Our performance experiments span \numberOfMethods{} methods, \numberOfDatasets{} datasets,
five budgets, five downstream models, and three seeds, yielding
79{,}875 runs,  totaling approximately 292 GPU days and 685 CPU days. Validity with 3 noise levels and stability results with 25 bootstrapped repetitions and respectively \numberOfMethods{} methods, \numberOfDatasets{} datasets and 5 repetitions took additional 2243 CPU days.

\subsection{Evaluation}\label{subsection-benchmarking-evaluation}
We assess the quality of a feature selection method in four dimensions: predictive performance, stability, validity, and runtime. For cardinality-constrained feature selection, we need to consider two hyperparameters that are independent of the specific method: the cardinality constraint $k$ and the time constraint. To obtain a diverse performance estimate across different application scenarios, we vary $k$ as a function of the number of features $p$. For datasets with $\leq 100$ features, we vary $k = \lceil p \times j/6 \rceil$, where $p$ is the number of features and $j \in \{1,\dots,5\}$ is an index variable. For bigger datasets, we use an exponential regime: $k = \lceil p^{j/6} \rceil$. For more details, see Appendix \ref{appendix-subsection-datasets-feature-budget}. If methods do not respect the cardinality constraint, we use random fallback (see Appendix \ref{par:fallback}). With respect to the time constraint, we use a limit of one hour per feature selection experiment consisting of a (dataset, budget, model)-tuple for all feature selection methods. 
The downstream model training and inference time is set to one hour per experiment as well.

\myparagraph{Predictive Performance.}\label{subsection-benchmarking-performance}
The predictive performance of a feature selection method is the performance of the downstream model using the selected feature subset on a given task. It is measured by ROC AUC for binary classification problems, log-loss for multi-class classification, and RMSE for regression. We discuss the choice of evaluation metrics in Appendix~\ref{par:evaluation-metrics}. To compare across these metrics, we report Elo ratings (Appendix~\ref{par:elo-rating}) and improvement over \textsc{Random}, $I(m,d) = 100 \cdot \bigl(\mathrm{Err}(r,d) -
\mathrm{Err}(m,d)\bigr) / \mathrm{Err}(r,d)$, where $\mathrm{Err}(m,d)$ is the metric error of method $m$ on dataset $d$ and $r$ denotes random selection (Appendix~\ref{par:ior}). We repeat all experiments for three seeds.

\myparagraph{Stability.}
For interpretability, feature selection methods should not only achieve strong predictive performance, but also yield similar subsets for similar data, for instance via resampling of the original data. We refer to this desideratum as \emph{stability}, which is sometimes also described as robustness, reliability, or continuity \citep{dembinsky2026unifying}. We assess the stability with the Nogueira metric \citep{nogueira2016stability} across 25 bootstrap samples. The Nogueira estimator corrects for chance, asymptotically assigns random selections a score of 0, and reaches a maximum of 1. Values below 0.40 indicate poor stability, values between 0.40 and 0.75 reflect intermediate to good stability, and values above 0.75 denote excellent stability \citep{nogueira2016stability}.

\myparagraph{Validity.}
In addition to being stable, feature selection methods should also recover relevant rather than irrelevant features, an interpretability desideratum that we refer to as \emph{validity}. In the literature, this property is also described as truthfulness, faithfulness, or fidelity \citep{dembinsky2026unifying}. However, assessing validity is challenging in the absence of a synthetic ground truth \citep{bakhmach2026roofs, bolon2013review, hopf2021benchmarkFilterFS}. 
We follow prior work \citep{cherepanova2023benchmarkFS, guyon2003design, guyon2004result, guyon2007benchmark} and evaluate validity by augmenting the original feature set with uninformative Gaussian noise variables at 50\%, 75\%, and 100\% of the original feature count. We then operationalize validity by quantifying the proportion of selected features from the original feature set, which we term selection precision. This measure ranges from 0 to 1. In our setting, the selection precision is equivalent to the selection recall, \textit{i.e.}, the proportion of original, informative features selected, when the recall is normalized by its maximum possible value for $k$.

\myparagraph{Runtime.}
Lastly, feature selection methods should produce well-performing, stable, and valid subsets of features as fast as possible. Therefore, we measure the wall-clock time a method needs to produce a feature subset for a given budget $k$ with the upper bound of one hour. Where the implementation allows it, exceeding the limit interrupts the method and triggers a random fallback that restores the subset to size $k$, either by adding randomly chosen unselected features or, for backward methods, by randomly subsampling the current set. To make runtimes comparable across datasets of different dimensionality, we normalize by the number of features and report seconds per 1{,}000 features.
\section{Results}\label{section-results}
\begin{figure}[ht]
  \centering
  \begin{subfigure}[t]{0.375\linewidth}
    \centering
    \includegraphics[width=\linewidth]{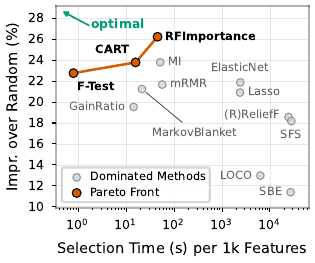}
  \end{subfigure}\hfill
  \begin{subfigure}[t]{0.605\linewidth}
    \centering
    \includegraphics[width=\linewidth]{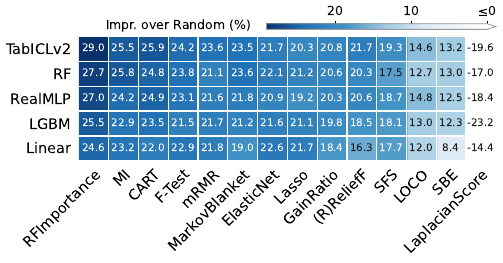}
  \end{subfigure}
  \caption {\textbf{Predictive performance.} Both panels report
improvement over \textsc{Random} (\%, higher is better), averaged over three seeds, five cardinalities, and \numberOfDatasets{} datasets. (\textbf{Left}): Improvement averaged over downstream models against feature selection time (s) per 1{,}000 features with Pareto front marked in orange. We exclude LaplacianScore as it performs worse than Random. (\textbf{Right}): Improvement for every model--method interaction.}
  \label{figure-results-tradeoffs}
\end{figure}

\begin{figure}[t]
  \centering
  \includegraphics[width=\textwidth]{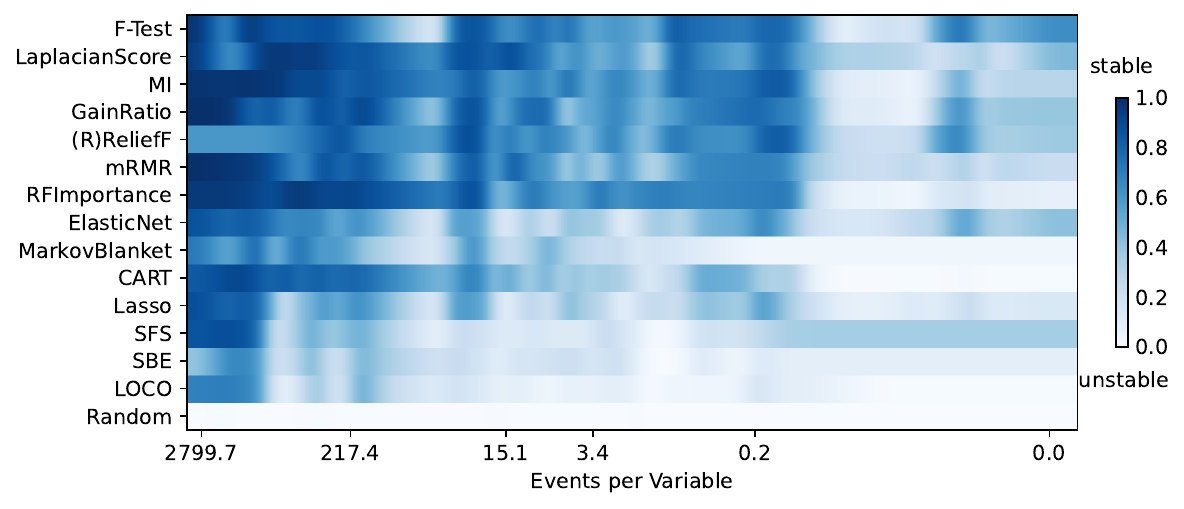}
  \caption{\textbf{Stability.} Nogueira score \citep{nogueira2016stability} across datasets. We use the events per variable, \textit{i.e.}, the number of samples per feature for regression problems or the number of samples in the minority class per feature for classification problems, as a proxy for the selection difficulty \citep{harrell2015regression, heinze2018variable}.}
  \label{figure-results-stability}
\end{figure}

\begin{figure}[h]
  \centering
  \includegraphics[width=\textwidth]{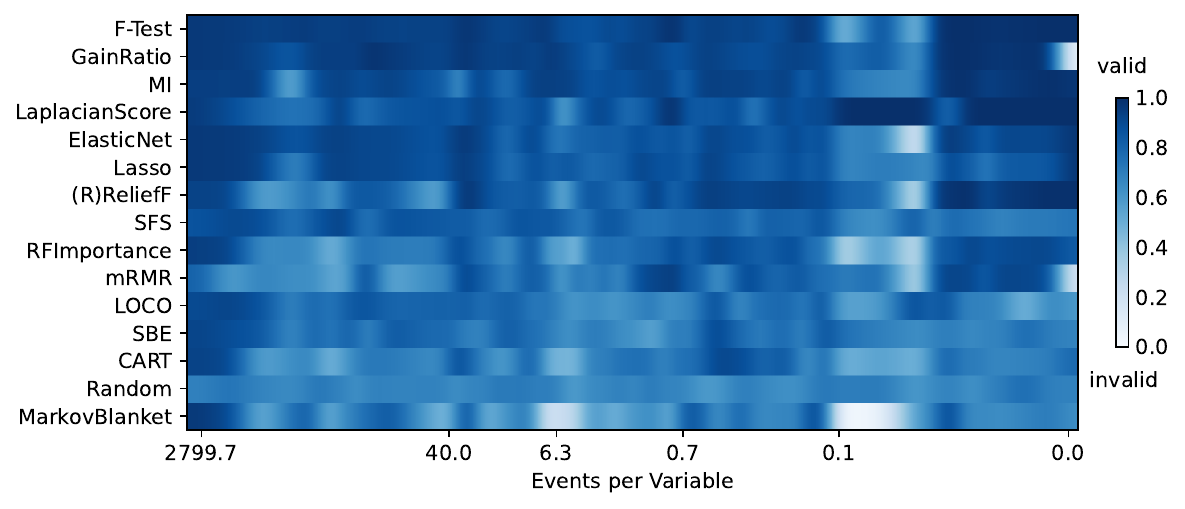}
  \caption{\textbf{Validity.} Selection precision across datasets at a noise level of 50\%. Other noise levels can be found in Appendix \ref{appendix-section-benchmarking}. Like for the stability, we use the events per variable, \textit{i.e.}, the number of samples per feature for regression problems or the number of samples in the minority class per feature for classification problems, as a proxy for the selection difficulty \citep{harrell2015regression, heinze2018variable}.}
  \label{figure-results-validity}
\end{figure}

We evaluated 15 widely used feature selection methods on 71 manually curated real-world tabular datasets with five cardinality constraints per dataset and five downstream predictors across three seeds in our performance experiments. The results are shown in Figures~\ref{figure-results_performance-elo} and \ref{figure-results-tradeoffs} and discussed further in the next paragraphs. For stability and validity runs, we use the same 15 methods and 71 datasets respectively and 5 repetitions. Each stability run consists of 25 bootstrapped repetition, each validity run includes 3 noise levels. The results are shown in Figures~\ref{figure-results-stability} and~\ref{figure-results-validity}, respectively, and discussed in the last paragraphs.

\myparagraph{Strong Methods Can Be Fast.}
The left plot in Figure~\ref{figure-results-tradeoffs} shows improvement over Random (\%) versus selection time (seconds per 1k features), with predictive performance and runtime on the axes; exact scores are in Table~\ref{table-feature-selection-performance}. The Pareto front consists of RFImportance and CART, the tree-based methods, together with the correlation-based F-Test. All three rank in the top four on predictive performance alone (MI ranks second) and in the six fastest on runtime alone (with LaplacianScore, GainRatio, and MarkovBlanket), so practitioners who value both face little trade-off. The search-based methods (LOCO, SBE, and SFS) sit at the opposite corner, performing poorly while being the slowest, although this, at least in part, can be attributed to the imposed time limit.

\myparagraph{Good Feature Selection Helps Every Model.}
The heatmap in Figure~\ref{figure-results-tradeoffs} presents improvement over Random (\%) for each combination of feature selection method and downstream model averaged over seeds, budgets, and datasets. We observe that model ordering is rather consistent across methods, with TabICLv2 ranking first for eight of the 14 methods and Linear last for seven. The improvement spread across models within a method (at most 5.4pp) is far smaller than the spread across methods within a model (15.8pp under TabICLv2). This asymmetry suggests that even the most expressive model, like TabICLv2, cannot recover predictive signal from a poorly selected subset and so the choice of feature selection method is crucial. We find no evidence that aligning the inductive biases of selector and predictor helps. Although Lasso and ElasticNet peak with the Linear model (but also only by 0.5pp), the tree-based RFImportance and CART achieve best performance with TabICLv2 rather than with RF or LGBM.

\myparagraph{Strong Methods are not Necessarily Stable.}\label{subsection-results-stability} The stability of the feature selection methods is shown in Figure \ref{figure-results-stability}. Most methods are stable when the number of samples (in the minority class) per feature is sufficiently large, but become increasingly unstable in low-data regimes \citep{harrell2015regression, heinze2018variable}. F-Test and distance-based methods exhibit the highest stability contributing to a better interpretability of the selected features, followed by information-theoretic methods. F-Test and MI are the only high-performing methods that rank high in stability. Due to its high fallback rates, search-methods are only slightly more stable than a random selection. An extended stability analysis can be found in Appendix \ref{appendix-subsection-benchmarking-stability}.

\myparagraph{Strong Methods do not Necessarily Select Informative Features.}\label{subsection-results-validity} Figure \ref{figure-results-validity} shows the validity of the feature selection methods. In contrast to the stability, the validity is less sensitive to the number of samples (in the minority class) per feature. F-Test and information-theoretic methods achieve the highest validity aiding interpretability of the selection, followed by distance-based methods. F-Test and MI are the only high-performing methods that rank high in validity. MarkovBlanket exhibits lower validity than even a random selection, potentially because it favors removing redundant features that are prevalent in real-world data \citep{guyon2003introduction, koller1996MarkovBlanket}. An extended validity analysis at different noise levels can be found in Appendix \ref{appendix-subsection-benchmarking-validity}.

\myparagraph{F-Test Is Strong on All Four Axes.}
Taken together, no single method dominates on all four evaluation axes, but among the strong candidates F-Test comes closest to being a well-balanced choice. It lies on the Pareto front of predictive performance and runtime, and ranks highest on both stability and validity (Figure~\ref{figure-results_performance-elo}). RFImportance and CART achieve higher Elo ratings but are less stable, and the model-coupled search methods are competitive on none of the four dimensions under our one-hour budget.
\section{Conclusions}\label{section-conclusions}
Feature selection is essential across domains where practitioners must balance predictive performance, computational constraints, and interpretability. 
However, existing benchmarks offer limited guidance to select methods that best suit different real-world requirements. 
To address this gap, we introduce \benchmarkName{}, a large-scale benchmark for feature selection on 71 manually curated real-world tabular datasets, spanning classification and regression tasks. 
We evaluate 15 widely used feature selection methods using standardized protocols and five downstream models, including a tabular foundation model. 
\benchmarkName{} is not only the largest general feature selection benchmark to date, but also the only one to include leading tabular models.

\myparagraph{Limitations.}\label{subsection-limitations}
While \benchmarkName{} substantially advances the state of feature selection benchmarking, several limitations remain. 
(1) All feature selection methods are evaluated under a one-hour time limit, which results in slower methods not always completing their task (up to 47\% of runs for SBE). 
Since we handle timeouts with a random fallback that keeps the feature budget fixed, the results for the affected methods reflect their performance under the limit on our hardware and are subject to change if given more compute.
Appendix~\ref{par:fallback} provides an ablation without the random fallback.
(2) Several methods are implemented by us in cases where no sufficiently maintained alternative was available, rather than taken from an established library (Table~\ref{table-benchmarking-selected-methods}). Although we follow the original algorithms, our implementations may be less optimized than library equivalents, which can disadvantage them under a wall-clock limit.
(3) Our analysis of stability and validity relies on bootstrapping and Gaussian noise, respectively. 
Both are arguably simpler than what methods face in practice, since bootstrapping resamples preserve the original feature correlations and Gaussian noise variables are independent of both the target and the informative features. 
Other forms of perturbation may affect feature selection behavior differently, \textit{e.g.}, non-Gaussian noise, outliers, missing values, distribution shifts, or class imbalances.
(4) All methods are constrained to return exactly $k$ features, however, some of them natively determine the subset size intrinsically, such as Lasso and ElasticNet via the regularization path or MarkovBlanket via conditional independence tests. 
Imposing a fixed cardinality makes them comparable but does not evaluate them as originally intended, and may understate their performance.
(5) The observed performance of feature selection methods strongly depends on the evaluation metrics used. 
Using the ROC AUC for binary classification, log-loss for multiclass classification, and RMSE for regression (Appendix~\ref{par:evaluation-metrics}), we employ widely adopted standard metrics, but alternative criteria such as calibration or fairness may lead to different results. 

\myparagraph{Recommendations.}\label{subsection-recommendations} 
Our results lead to several practical recommendations: 
(1) When predictive performance is the primary objective, practitioners might want to choose a tree-based method such as RFImportance or CART. These methods are also strong candidates when runtime is considered.
(2) For interpretability-focused applications, F-Test emerges as the best choice. 
It ranks highest in terms of stability and validity, next to information-theory and distance-based methods, with a competitive predictive performance and runtime. 
(3) The choice of a suitable downstream model for a feature selection method can be important. A state-of-the-art model like TabICLv2 is mostly, but not always, a better predictor than RF, LGBM or even a linear model. Therefore, practitioners should evaluate whether simpler downstream models may be preferable to more complex and computationally expensive alternatives.

\newpage
\newpage

\section*{Reproducibility statement}
SelectArena is an open-source framework. We publish the code and our curated datasets. More information can be found in Appendix \ref{appendix-subsection-benchmarking-links}. In the Benchmarking Setup (Chapter~\ref{subsection-benchmarking-hardware}), we detail the hardware setup used for the experiments. With fully providing soft- and hardware details as well as the datasets used, SelectArena is fully reproducible.

\section*{Acknowledgment Statement}
We thank the dataset donors for sharing their data, especially those under a public license. 
Furthermore, we thank the community, particularly OpenML, Kaggle, and Hugging Face, for developing tools that facilitate the sharing of tabular data. 
We thank the author of the scikit feature library for the inspiration through the provided implementations of feature selection methods.
B. Schäfer, A Liefooghe and A. Sokolov have benefited from the support of the National Research Agency under France 2030, MAIA Project ANR-22-EXES-0009, and from the "Pôle Métropolitain Côte d'Opale" (PMCO).
D. Matus and L. Purucker acknowledge funding by the Deutsche Forschungsgemeinschaft (DFG, German Research
Foundation) through SFB 1597 (SmallData), grant number 499552394, and through EFRE (FEIH\_2698644) grant number 539134284, as well as the state of Baden-Württemberg. F. Hutter acknowledges the financial support of the Hector Foundation.
R. Knauer and E. Rodner acknowledge being supported by the European Union's Erasmus+ program through EUonAIR, reference 101177370, and the Deutsche Forschungsgemeinschaft (DFG) through Berlin Initiative for Applied Foundation Model Research, reference 528483508.

\section*{Author Contributions}
B. Schäfer co-led the code development, led the method curation, co-led the dataset curation, contributed the method implementation, co-led the experiments, co-led the evaluation and the visualization and co-led the writing of the paper. 
D. Matus co-led the code development, contributed to the method curation, co-led the dataset curation, led the method implementation, co-led the experiments, co-led the evaluation and the visualization and co-led the writing of the paper. 
L. Purucker co-led the code development, supervised the method curation, supervised the dataset curation, supervised the experiments and supervised the writing of the paper. 
R. Knauer supervised the project and contributed to writing and reviewing the paper. 
E. Rodner supervised R. Knauer and helped with reviewing and editing the paper. 
F. Hutter supervised L. Purucker and D. Matus. 
A. Sokolov supervised B. Schäfer. 
A. Liefooghe supervised B. Schäfer and helped with reviewing and editing the paper.

\section*{Competing Interests}
L. Purucker and F. Hutter are a subset of the authors of TabArena. The authors declare no other competing interests.

\newpage

\bibliographystyle{iclr2027_conference}
\bibliography{lib,proc,strings,localbib}


\newpage
\appendix

\begin{appendices}
\listofappendices
\counterwithin{figure}{section}
\counterwithin{table}{section}
\crefalias{section}{appendix}
\crefalias{subsection}{appendix}
\newpage
\input{chapters/7_appendix}
\end{appendices}

\end{document}